\def\PUBLICVERSION{1}
\documentclass[11pt]{article}
\ifdefined\PUBLICVERSION
  \usepackage[preprint]{acl}
\else
  \usepackage[review]{acl}
\fi
\usepackage{times,latexsym}
\usepackage[T1]{fontenc}
\usepackage[utf8]{inputenc}
\usepackage{microtype,inconsolata}
\usepackage{url,booktabs,amsmath,graphicx,xcolor,tcolorbox,float,stfloats,placeins}
\usepackage{eso-pic,balance}
\AddToShipoutPictureFG{\AtPageLowerLeft{\put(72,28){\normalfont\fontsize{9}{11}\selectfont\color{gray}Work in progress}}}

\usepackage{amsmath,amsfonts,bm}

\def\eqref#1{equation~\ref{#1}}

\def\1{\bm{1}}

\DeclareMathAlphabet{\mathsfit}{\encodingdefault}{\sfdefault}{m}{sl}
\SetMathAlphabet{\mathsfit}{bold}{\encodingdefault}{\sfdefault}{bx}{n}

\newif\ifpublicversion
\ifdefined\PUBLICVERSION\publicversiontrue\else\publicversionfalse\fi
\title{Long-Horizon RL Needs Muon}
\ifpublicversion
  \author{%
Kai Ruan\textsuperscript{1,*}\quad Jinghao Lin\textsuperscript{*}\quad Tao Chen\textsuperscript{*}\\
\textbf{Ziqi Zhou\textsuperscript{2}\quad Qianshan Wei\textsuperscript{3}\quad Zihe Huang\textsuperscript{4}}\\[0.5em]
{\small \textsuperscript{1}Gaoling School of Artificial Intelligence, Renmin University of China}\\
{\small \textsuperscript{2}Duke University}\\
{\small \textsuperscript{3}Institute of Automation, Chinese Academy of Sciences}\\
{\small \textsuperscript{4}Institute of Computing Technology, Chinese Academy of Sciences}\\[0.3em]
{\small \textsuperscript{*}Equal contribution.}
}

\else
  \author{Anonymous authors\\Paper under double-blind review}
\fi
\begin{document}
\raggedbottom
\maketitle
\begin{abstract}
Long-horizon agents must coordinate reasoning, tool use, and environment feedback over extended interactions. Learning from sparse outcome rewards requires weighting trajectories of different lengths and updating the policy while rollouts are still being collected. We examine a single-rollout asynchronous recipe combining Muon, token-averaged loss, truncated importance sampling, and auxiliary length and tool-error penalties. Single-rollout sampling allows asynchronous learning without waiting for multiple trajectories for the same issue. We study optimizer choices and recipe components for repository-level issue resolution across two model families. The evaluated training runs show higher training success with the full recipe than with the baselines.

\end{abstract}

\begin{figure*}[!b]
\centering
\includegraphics[width=\textwidth]{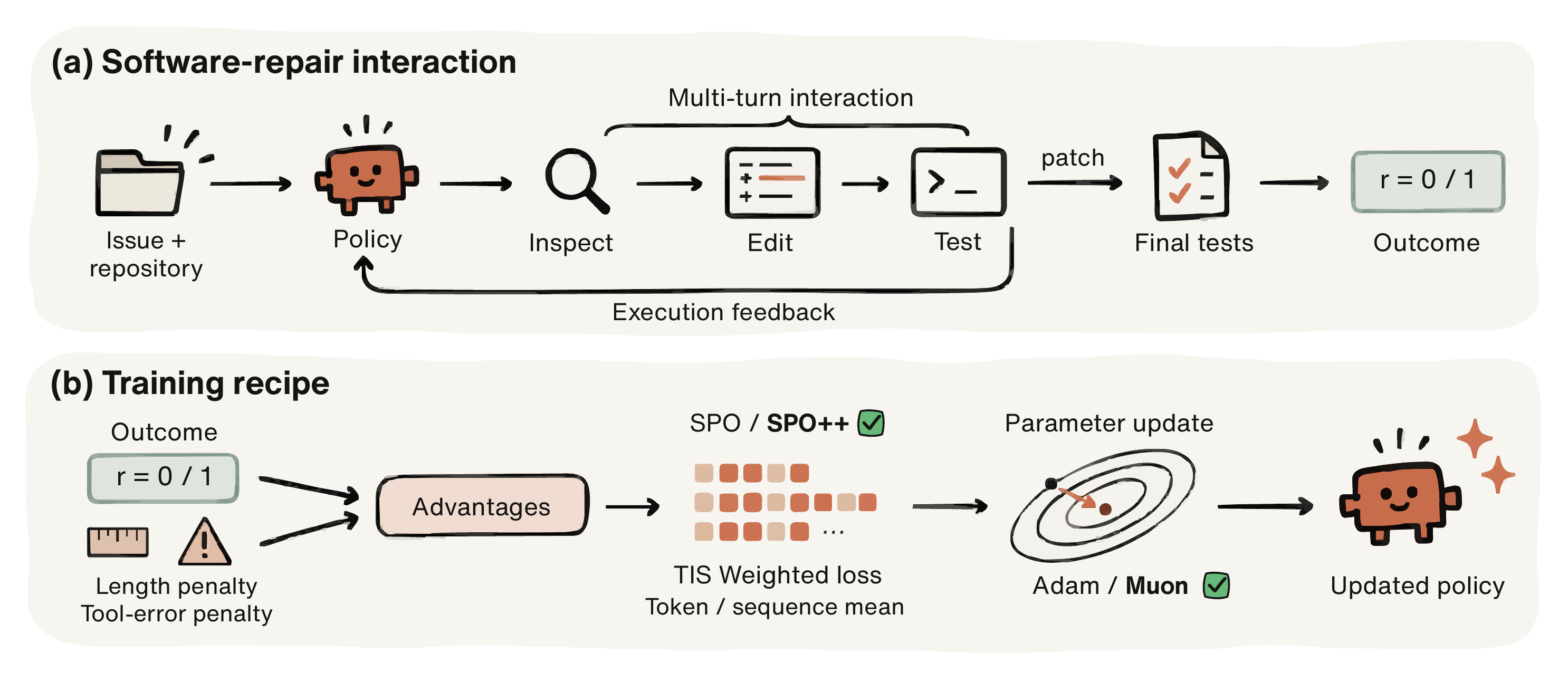}
\caption{Overview of the training recipe. (a) A software engineering agent inspects, edits, and tests a repository; final tests provide the binary outcome reward. (b) Length and tool-error penalties shape advantages, TIS reweights token-level policy losses, and token-level averaging aggregates the policy loss before the Muon update. }
\label{fig:teaser}
\end{figure*}

\section{Introduction}
Repository-level issue resolution requires an agent to interpret an issue description, navigate and edit the codebase, and validate a candidate patch \citep{jimenez2023swebench,yang2024sweagent}. Such workflows require coordinated execution across dependent steps, a defining feature of long-horizon agents \citep{wu2026eveai4ai}. Executable tests make these interactions a source of reinforcement-learning feedback \citep{jain2025r2egym,luo2025deepswe,song2026swemaster}. Trajectory lengths vary, and asynchronous collection introduces a gap between rollout and learner policies. We study how optimizer choice affects learning from sparse outcomes under these conditions.

We compare Adam \citep{kingma2014adam} and Muon \citep{jordan2024muon} within Single-stream Policy Optimization (SPO) \citep{xu2025spo}, then study how Muon combines with SPO++ \citep{ruan2026spoplusplus}, reward shaping, and importance correction. Ablations examine the length penalty \citep{yu2025dapo}, tool-error penalty, and update rules using the same training success rate metric.

\begin{figure}[t]
\centering
\includegraphics[width=\linewidth]{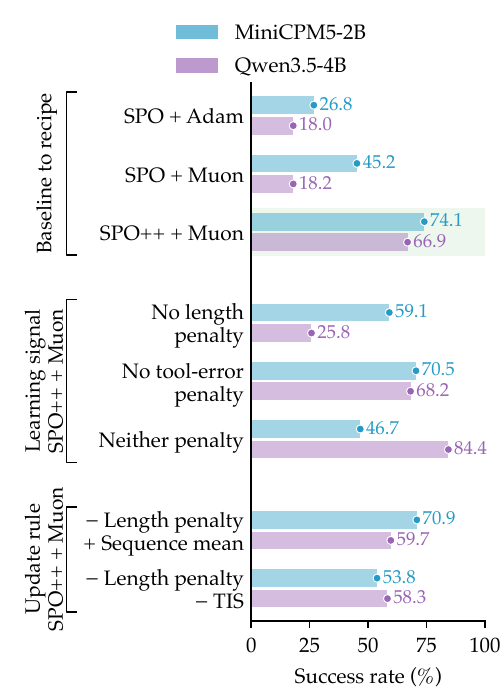}
\caption{Recipe comparisons for the selected training runs in Figures~\ref{fig:recipe-reward-minicpm} and~\ref{fig:recipe-reward-qwen}. Bars summarize unsmoothed training success rates. Pale green highlights the full recipe.}
\label{fig:story-components}
\end{figure}

The full recipe achieves higher training success than the SPO baselines in the selected runs from both model families (Figure~\ref{fig:story-components}).

\section{Related Work}
\label{sec:related-work}
\paragraph{Learning from agent--environment interaction.}
Software engineering (SWE) agents interact with code repositories and use execution-based evaluation \citep{jimenez2023swebench,yang2024sweagent,wang2024openhands}, supported by environments and training tasks such as SWE-Gym, R2E-Gym, and SWE-smith \citep{pan2024swegym,jain2025r2egym,yang2025swesmith,lin2026codingagentdata}. Training feedback can come from software evolution \citep{wei2025swerl}, test execution \citep{luo2025deepswe,song2026swemaster}, or simulation \citep{sun2026sweworld}. Our experiments use execution-based rewards to study policy optimization.

\paragraph{Single-rollout and asynchronous optimization.}
SPO and SPO++ provide the single-rollout foundation of our recipe \citep{xu2025spo,ruan2026spoplusplus}. SAO uses value-model learning and token-level masking based on importance ratios \citep{hou2026sao}, whereas FlashReinforce combines critic-free updates with token importance sampling and a sequence-level trust region \citep{hu2026flashreinforce}. Other methods address off-policy agent training through turn-level correction \citep{li2025sorl} and variance control \citep{huang2026vcpo}. Muon has also been studied in mathematical and interactive RL \citep{fan2026pion,ruan2026muonagentic}. We study Muon within single-rollout asynchronous training for issue resolution.

\paragraph{Reward and loss design.}
Reward shaping determines how task outcomes and intermediate behavior contribute to learning, while loss aggregation determines the weight of trajectories with different lengths. Both are studied in agentic RL recipes \citep{wang2026arlarena,wu2026agentstar,masud2026rewardengineering} and in reasoning RL through length penalties and trajectory weighting \citep{yu2025dapo,zeng2026balanced}. We examine their combination with Muon in multi-turn software repair.

\begin{figure*}[!t]
\centering
\includegraphics[width=\linewidth]{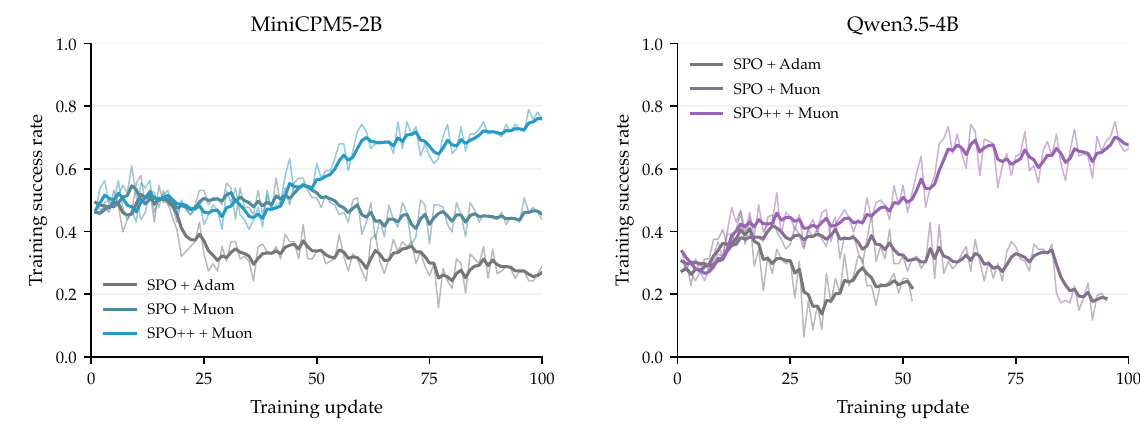}
\caption{Selected training runs. Thin translucent lines show raw values; bold lines show exponential moving averages (span 5) from the first observation.}
\label{fig:story-baselines}
\end{figure*}

\section{Method}
\label{sec:method}
\subsection{Muon-based policy optimization}
For each sampled issue, the agent produces one rollout by interacting with the repository through tools. Rollout collection proceeds asynchronously with policy updates, introducing policy lag.

Muon's update direction admits a spectral-norm interpretation \citep{bernstein2025deriving}. For a nonzero matrix gradient $G$ with compact SVD $G=U_r\Sigma_r V_r^\top$, one solution to the local linear problem is
\begin{equation}
 \begin{aligned}
 \min_{\|\Delta W\|_2\leq\eta}\;\langle G,\Delta W\rangle,
 \qquad \Delta W^*=-\eta U_rV_r^\top,
 \end{aligned}
 \label{eq:muon-geometry}
\end{equation}
where $\|\cdot\|_2$ denotes the spectral norm and $\eta>0$ is the update budget. This direction preserves the gradient's singular vectors and equalizes its nonzero singular values. The constraint bounds the change in a linear layer's output: $\|\Delta W x\|_2\leq\eta\|x\|_2$. Practical Muon applies approximate orthogonalization to a momentum-based update, with matrix-dependent scaling.

Newton--Schulz iterations compute this approximation using matrix multiplications, avoiding an explicit SVD \citep{jordan2024muon}. Our implementation applies it to hidden weight matrices; embeddings, the output head, normalization parameters, and vector parameters use AdamW.

The SPO baseline computes advantages from outcome rewards and its task-specific baseline, then standardizes them across trajectories and uses proximal policy optimization (PPO)-style clipping \citep{schulman2017ppo,xu2025spo}. We define valid tokens as agent-generated tokens included in the policy loss. Our recipe follows SPO++ in standardizing advantages across valid tokens \citep{ruan2026spoplusplus}, with auxiliary reward shaping for issue resolution (Figure~\ref{fig:teaser}).

\label{sec:supplementary}
\label{sec:recipes}
Following DAPO's Clip-Higher settings \citep{yu2025dapo}, the PPO importance-sampling ratio uses asymmetric clipping with $\epsilon_{\mathrm{low}}=0.2$ and $\epsilon_{\mathrm{high}}=0.28$.

\subsection{Outcome reward and auxiliary penalties}
\label{sec:penalty-definition}
Following SWE-bench \citep{jimenez2023swebench}, the final patch receives a binary outcome reward of 1 if it passes all fail-to-pass (F2P) and pass-to-pass (P2P) tests, and 0 otherwise. F2P tests check issue resolution; P2P tests check for regressions. Empty test sets and evaluations interrupted by infrastructure failures remain unscored.

The length penalty discourages long agent responses at each interaction turn. The tool-error penalty applies to invalid actions and designated tool, command, session, or execution errors reported in observations.

When triggered, the tool-error term is $-0.5$. The length penalty reaches $-0.5$ for responses exceeding 2,048 tokens, with the trajectory total proportionally capped. We omit this penalty on the final turn contributing to the policy loss. Writing the length and tool-error terms at turn $k$ as $c_k^{\mathrm{len}}$ and $c_k^{\mathrm{tool}}$, their combined contribution is
\begin{equation}
 c_k=\operatorname{clip}\!\left(c_k^{\mathrm{len}}+c_k^{\mathrm{tool}},-0.25,0.25\right).
 \label{eq:turn-penalty}
\end{equation}
The combined term is added to each valid token's advantage in that turn, before standardization across valid tokens. Agent responses containing predefined command-rejection messages trigger an additional trajectory-level penalty. Penalty ablations leave the tool interface and binary outcome reward unchanged.

\subsection{Importance correction and loss aggregation}
\label{sec:tis-definition}
Truncated importance sampling (TIS) reweights tokens to account for differences between training and sampling policies \citep{yao2025offpolicy}, arising from execution backends and asynchronous policy lag \citep{zheng2025stabilizing}. With frozen pre-update policy $\pi_{\mathrm{old}}^{\mathrm{train}}$ evaluated by the training backend and sampling policy $\pi^{\mathrm{rollout}}$, the weight for generated token $a_t$ in context $s_t$ is
\begin{equation}
 w_t=\min\!\left(C,\frac{\pi_{\mathrm{old}}^{\mathrm{train}}(a_t\mid s_t)}{\pi^{\mathrm{rollout}}(a_t\mid s_t)}\right).
\end{equation}
We use $C=2$ and apply the TIS weights directly to the masked token-level PPO losses, without batchwise weight normalization. The no-TIS ablation removes the weights while retaining PPO ratio clipping.

Loss aggregation determines how trajectories of different lengths contribute to an update \citep{liu2025drgrpo,yu2025dapo,zeng2026balanced}. For a batch of $N$ trajectories, let $T_i>0$ count the valid tokens in trajectory $i$ and $\ell_{it}$ denote their corrected losses. Tokens from environment observations are excluded from the policy loss. We compare
\begin{align}
L_{\mathrm{token}} &= \frac{\sum_{i=1}^{N}\sum_{t=1}^{T_i}\ell_{it}}{\sum_{i=1}^{N} T_i},\\
L_{\mathrm{sequence}} &= \frac{1}{N}\sum_{i=1}^{N}\frac{1}{T_i}\sum_{t=1}^{T_i}\ell_{it}.
\end{align}
The main recipe uses token-level averaging, which weights each trajectory mean in proportion to $T_i$. The sequence-level averaging ablation weights trajectory means equally and keeps advantage standardization unchanged.

\begin{figure*}[!tbp]
\centering
\includegraphics[width=\linewidth,height=0.84\textheight,keepaspectratio]{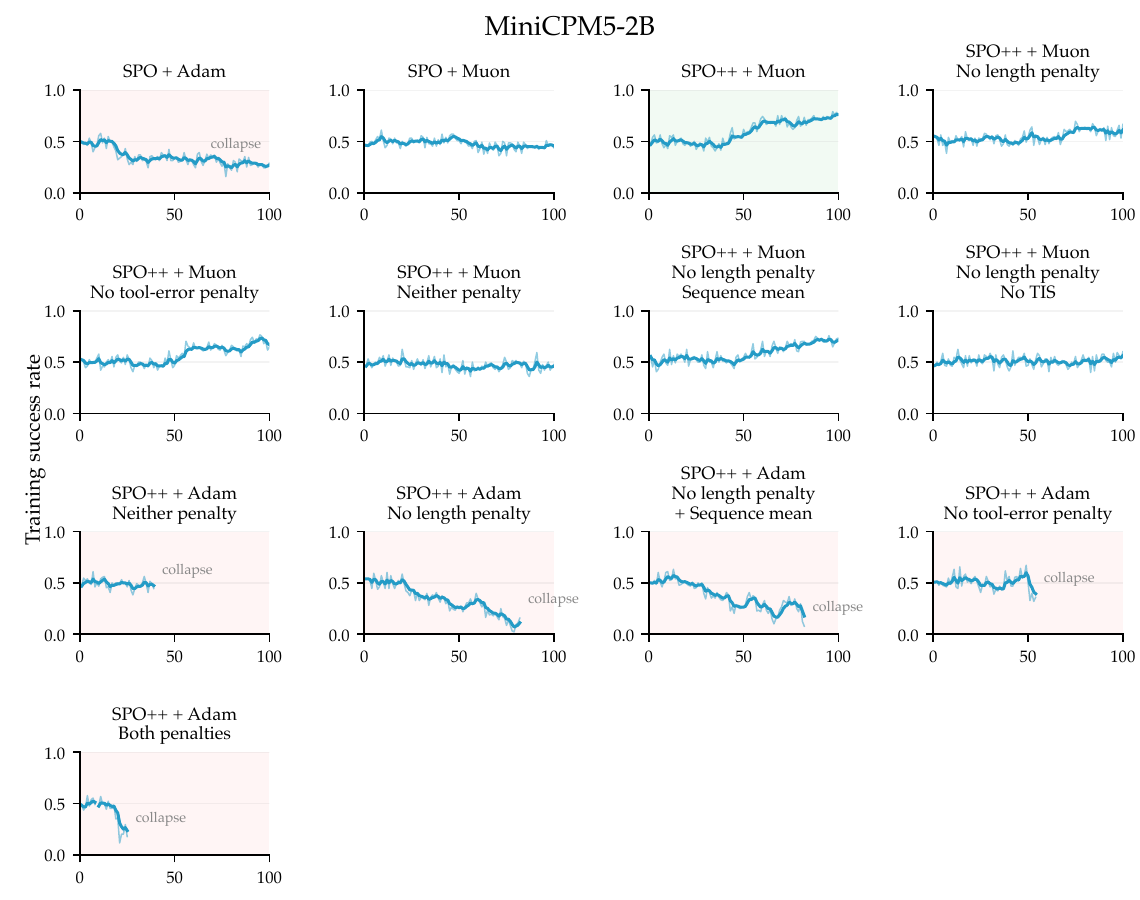}
\caption{Training success across MiniCPM5-2B recipes: thin lines show unsmoothed success rates, bold lines show span-5 EMA, red marks collapse, green marks the full recipe.}
\label{fig:recipe-reward-minicpm}
\end{figure*}

\begin{figure*}[!tbp]
\centering
\includegraphics[width=\linewidth,height=0.84\textheight,keepaspectratio]{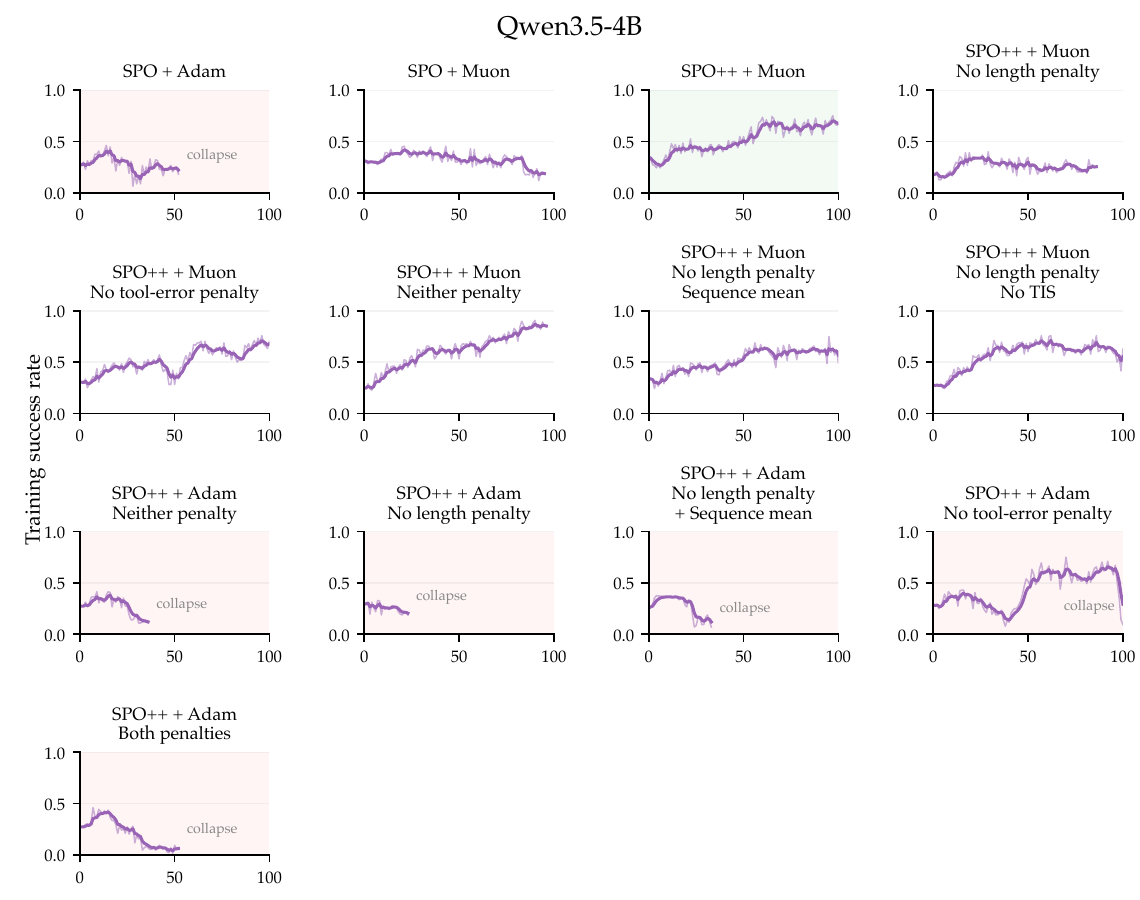}
\caption{Training success across Qwen3.5-4B recipes: thin lines show unsmoothed success rates, bold lines show span-5 EMA, red marks collapse, green marks the full recipe.}
\label{fig:recipe-reward-qwen}
\end{figure*}

\section{Experiments}
\label{sec:experiments}
\subsection{Experimental setup}
\label{sec:protocol}
Training uses issue-resolution tasks from R2E-Gym \citep{jain2025r2egym}. The agent interacts with the execution environment through the Claude Code harness (version 2.1.118). Training success rate is the fraction of evaluated rollouts whose final patches pass all required tests in each update; auxiliary penalties are excluded.

We evaluate MiniCPM5-2B \citep{openbmb2026minicpm5} and Qwen3.5-4B \citep{qwen2026qwen35} with the same training-update and rollout budgets across independently initialized runs.

Curves show the longest available training run per configuration, with a separately selected run for MiniCPM5-2B with SPO + Muon (Figure~\ref{fig:story-baselines}). Figure~\ref{fig:story-components} summarizes the same runs using mean success over updates 90--100 when that window is complete, and the last observed value otherwise. Numerical summaries use recorded values; matched comparisons use complete windows from runs with the same seed.

Adam and Muon share the SPO baseline's data, rollout budgets, termination rules, and loss aggregation. Both collect task-specific outcome statistics online, without SPO's offline initialization.

The full-recipe comparison holds Muon fixed while jointly changing advantage normalization, task-specific outcome estimation, clipping, and auxiliary penalties. Component ablations remove either or both penalties, or vary TIS and loss aggregation with the length penalty disabled. Each configuration collects its own trajectories through asynchronous rollouts; comparisons pair runs with the same seed.

\subsection{Main results}
\label{sec:design}

The full recipe reaches higher success rates than both SPO baselines later in training for both model families (Figure~\ref{fig:story-baselines}). The gain over SPO + Muon supports combining the optimizer with the full set of recipe changes.

Across the recipe comparisons, training success continues to improve with the full recipe and declines in several Adam configurations (Figures~\ref{fig:recipe-reward-minicpm} and~\ref{fig:recipe-reward-qwen}). Without the length penalty, the relative performance of token-level and sequence-level averaging varies across seeds.

\FloatBarrier
\section{Conclusion}
Combining Muon with single-rollout asynchronous updates, importance correction, and auxiliary reward shaping yields higher training success than the SPO baselines in the evaluated runs from both model families.

\balance


\begin{thebibliography}{34}
\providecommand{\natexlab}[1]{#1}

\bibitem[{Bernstein(2025)}]{bernstein2025deriving}
Jeremy Bernstein. 2025.
\newblock \href {https://jeremybernste.in/writing/deriving-muon} {Deriving
  muon}.

\bibitem[{Fan et~al.(2026)Fan, Liu, Hong, Kompella, and Liu}]{fan2026pion}
Chongyu Fan, Gaowen Liu, Mingyi Hong, Ramana~Rao Kompella, and Sijia Liu. 2026.
\newblock \href {https://doi.org/10.48550/arXiv.2605.19282} {{Rethinking {Muon}
  Beyond Pretraining: Spectral Failures and High-Pass Remedies for {VLA} and
  {RLVR}}}.
\newblock \emph{Preprint}, arXiv:2605.19282.
\newblock Version v1, 2026-05-19.

\bibitem[{Hou et~al.(2026)Hou, Li, Tang, and Dong}]{hou2026sao}
Zhenyu Hou, Yujiang Li, Jie Tang, and Yuxiao Dong. 2026.
\newblock \href {https://doi.org/10.48550/arXiv.2607.07508} {Single-rollout
  asynchronous optimization for agentic reinforcement learning}.
\newblock \emph{Preprint}, arXiv:2607.07508.
\newblock Version v1.

\bibitem[{Hu et~al.(2026)Hu, Zhang, Zhang, Xu, Zhang, Peng, Yu, Molchanov,
  Kautz, and Dong}]{hu2026flashreinforce}
Jian Hu, Yifan Zhang, Hao Zhang, Binfeng Xu, Shaokun Zhang, Hongqing Peng,
  Zhiding Yu, Pavlo Molchanov, Jan Kautz, and Yi~Dong. 2026.
\newblock \href
  {https://yifanzhang-pro.github.io/FlashREINFORCE/FlashREINFORCE.pdf}
  {{FlashREINFORCE}: Critic-free single-rollout asynchronous {RL} for agentic
  language models}.

\bibitem[{Huang et~al.(2026)Huang, Zhang, Hu, Yang, and Han}]{huang2026vcpo}
Luke~J. Huang, Zhuoyang Zhang, Qinghao Hu, Shang Yang, and Song Han. 2026.
\newblock \href {https://arxiv.org/abs/2602.17616} {{Stable Asynchrony:
  Variance-Controlled Off-Policy RL for LLMs}}.
\newblock \emph{Preprint}, arXiv:2602.17616.

\bibitem[{Jain et~al.(2025)Jain, Singh, Shetty, Zheng, Sen, and
  Stoica}]{jain2025r2egym}
Naman Jain, Jaskirat Singh, Manish Shetty, Liang Zheng, Koushik Sen, and Ion
  Stoica. 2025.
\newblock \href {https://arxiv.org/abs/2504.07164} {{R2E-Gym}: Procedural
  environments and hybrid verifiers for scaling open-weights {SWE} agents}.
\newblock \emph{Preprint}, arXiv:2504.07164.
\newblock Version 1, 9 April 2025.

\bibitem[{Jimenez et~al.(2023)Jimenez, Yang, Wettig, Yao, Pei, Press, and
  Narasimhan}]{jimenez2023swebench}
Carlos~E. Jimenez, John Yang, Alexander Wettig, Shunyu Yao, Kexin Pei, Ofir
  Press, and Karthik Narasimhan. 2023.
\newblock \href {https://doi.org/10.48550/arXiv.2310.06770} {{SWE}-bench: Can
  language models resolve real-world {GitHub} issues?}
\newblock \emph{Preprint}, arXiv:2310.06770.
\newblock Version v3.

\bibitem[{{JLULLM LAB}(2026)}]{lin2026codingagentdata}
{JLULLM LAB}. 2026.
\newblock \href {https://linjh1118.github.io/coding-agent-data/} {From issues
  to learning signals: A survey of data for coding agents}.
\newblock Survey manuscript.

\bibitem[{Jordan et~al.(2024)Jordan, Jin, Boza, You, Cesista, Newhouse, and
  Bernstein}]{jordan2024muon}
Keller Jordan, Yuchen Jin, Vlado Boza, Jiacheng You, Franz Cesista, Laker
  Newhouse, and Jeremy Bernstein. 2024.
\newblock \href {https://kellerjordan.github.io/posts/muon/} {{Muon}: An
  optimizer for hidden layers in neural networks}.
\newblock Technical blog.

\bibitem[{Kingma and Ba(2014)}]{kingma2014adam}
Diederik~P. Kingma and Jimmy Ba. 2014.
\newblock \href {https://arxiv.org/abs/1412.6980} {{Adam}: A method for
  stochastic optimization}.
\newblock \emph{Preprint}, arXiv:1412.6980.

\bibitem[{Li et~al.(2025)Li, Elmahdy, Boyd, Wang, Zeng, Garcia, Bhatia,
  Kass-Hout, Xiao, and Hong}]{li2025sorl}
Chenliang Li, Adel Elmahdy, Alex Boyd, Zhongruo Wang, Siliang Zeng, Alfredo
  Garcia, Parminder Bhatia, Taha Kass-Hout, Cao Xiao, and Mingyi Hong. 2025.
\newblock \href {https://arxiv.org/abs/2511.20718} {{Stabilizing Off-Policy
  Training for Long-Horizon LLM Agent via Turn-Level Importance Sampling and
  Clipping-Triggered Normalization}}.
\newblock \emph{Preprint}, arXiv:2511.20718.

\bibitem[{Liu et~al.(2025)Liu, Chen, Li, Qi, Pang, Du, Lee, and
  Lin}]{liu2025drgrpo}
Zichen Liu, Changyu Chen, Wenjun Li, Penghui Qi, Tianyu Pang, Chao Du, Wee~Sun
  Lee, and Min Lin. 2025.
\newblock \href {https://arxiv.org/abs/2503.20783} {{Understanding R1-Zero-Like
  Training: A Critical Perspective}}.
\newblock \emph{Preprint}, arXiv:2503.20783.

\bibitem[{Luo et~al.(2025)Luo, Jain, Singh, Tan, Patel, Wu, Ariyak, Cai,
  Venkat, Zhu, Athiwaratkun, Roongta, Zhang, Li, Popa, Sen, and
  Stoica}]{luo2025deepswe}
Michael Luo, Naman Jain, Jaskirat Singh, Sijun Tan, Ameen Patel, Qingyang Wu,
  Alpay Ariyak, Colin Cai, Tarun Venkat, Shang Zhu, Ben Athiwaratkun, Manan
  Roongta, Ce~Zhang, Li~Erran Li, Raluca~Ada Popa, Koushik Sen, and Ion Stoica.
  2025.
\newblock \href {https://www.together.ai/blog/deepswe} {{DeepSWE}: Training a
  fully open-sourced, state-of-the-art coding agent by scaling {RL}}.
\newblock Together AI technical blog.
\newblock Published 2 July 2025; accessed 23 September 2026.

\bibitem[{Masud et~al.(2026)Masud, Wasi, Rahman, and
  Parvez}]{masud2026rewardengineering}
Md~Rayhanul Masud, Azmine~Toushik Wasi, Salman Rahman, and Md~Rizwan Parvez.
  2026.
\newblock \href {https://arxiv.org/abs/2601.19100} {Reward engineering for
  software tasks: A survey of reinforcement learning approaches}.
\newblock \emph{Preprint}, arXiv:2601.19100.

\bibitem[{{OpenBMB}(2026)}]{openbmb2026minicpm5}
{OpenBMB}. 2026.
\newblock \href {https://huggingface.co/openbmb/MiniCPM5-2B} {{MiniCPM5-2B}
  model card}.
\newblock Hugging Face model card.

\bibitem[{Pan et~al.(2024)Pan, Wang, Neubig, Jaitly, Ji, Suhr, and
  Zhang}]{pan2024swegym}
Jiayi Pan, Xingyao Wang, Graham Neubig, Navdeep Jaitly, Heng Ji, Alane Suhr,
  and Yizhe Zhang. 2024.
\newblock \href {https://doi.org/10.48550/arXiv.2412.21139} {Training software
  engineering agents and verifiers with {SWE}-gym}.
\newblock \emph{Preprint}, arXiv:2412.21139.

\bibitem[{{Qwen Team}(2026)}]{qwen2026qwen35}
{Qwen Team}. 2026.
\newblock \href {https://qwen.ai/blog?id=qwen3.5} {{Qwen3.5}: Towards native
  multimodal agents}.

\bibitem[{Ruan et~al.(2026{\natexlab{a}})Ruan, Lin, Huang, Zhou, Wei, Wang, and
  Sun}]{ruan2026muonagentic}
Kai Ruan, Jinghao Lin, Zihe Huang, Ziqi Zhou, Qianshan Wei, Xuan Wang, and Hao
  Sun. 2026{\natexlab{a}}.
\newblock \href {https://doi.org/10.48550/arXiv.2607.16169} {{When Does Muon
  Help Agentic Reinforcement Learning?}}
\newblock \emph{Preprint}, arXiv:2607.16169.
\newblock Version v4, 2026-08-02.

\bibitem[{Ruan et~al.(2026{\natexlab{b}})Ruan, Lin, Wei, Zhou, and
  Huang}]{ruan2026spoplusplus}
Kai Ruan, Jinghao Lin, Qianshan Wei, Ziqi Zhou, and Zihe Huang.
  2026{\natexlab{b}}.
\newblock \href {https://doi.org/10.48550/arXiv.2608.24870} {{SPO++:
  Stream-Aligned Policy Optimization for Asynchronous Agentic RL}}.
\newblock \emph{Preprint}, arXiv:2608.24870.
\newblock Version v1, 2026-08-25.

\bibitem[{Schulman et~al.(2017)Schulman, Wolski, Dhariwal, Radford, and
  Klimov}]{schulman2017ppo}
John Schulman, Filip Wolski, Prafulla Dhariwal, Alec Radford, and Oleg Klimov.
  2017.
\newblock \href {https://arxiv.org/abs/1707.06347} {Proximal policy
  optimization algorithms}.
\newblock \emph{Preprint}, arXiv:1707.06347.

\bibitem[{Song et~al.(2026)Song, Huang, Sun, Jiang, Le, Cheng, Chen, Hu, Chen,
  Jia, Zhao, Song, Zhang, and Wen}]{song2026swemaster}
Huatong Song, Lisheng Huang, Shuang Sun, Jinhao Jiang, Ran Le, Daixuan Cheng,
  Guoxin Chen, Yiwen Hu, Zongchao Chen, Yiming Jia, Wayne~Xin Zhao, Yang Song,
  Tao Zhang, and Ji-Rong Wen. 2026.
\newblock \href {https://arxiv.org/abs/2602.03411} {{SWE-Master: Unleashing the
  Potential of Software Engineering Agents via Post-Training}}.
\newblock \emph{Preprint}, arXiv:2602.03411.

\bibitem[{Sun et~al.(2026)Sun, Song, Huang, Jiang, Le, Lv, Chen, Hu, Luo, Zhao,
  Song, Xu, Zhang, and Wen}]{sun2026sweworld}
Shuang Sun, Huatong Song, Lisheng Huang, Jinhao Jiang, Ran Le, Zhihao Lv,
  Zongchao Chen, Yiwen Hu, Wenyang Luo, Wayne~Xin Zhao, Yang Song, Hongteng Xu,
  Tao Zhang, and Ji-Rong Wen. 2026.
\newblock \href {https://arxiv.org/abs/2602.03419} {{SWE-World: Building
  Software Engineering Agents in Docker-Free Environments}}.
\newblock \emph{Preprint}, arXiv:2602.03419.

\bibitem[{Wang et~al.(2026)Wang, Zhang, Wang, Shi, Li, Han, Tong, Deng, Taylor,
  Sun, Zhu, Cong, Sun, and Wang}]{wang2026arlarena}
Xiaoxuan Wang, Han Zhang, Haixin Wang, Yidan Shi, Ruoyan Li, Kaiqiao Han,
  Chenyi Tong, Haoran Deng, Alexander~K. Taylor, Renliang Sun, Yanqiao Zhu,
  Jason Cong, Yizhou Sun, and Wei Wang. 2026.
\newblock \href {https://arxiv.org/abs/2602.21534} {Arlarena: A unified
  framework for stable agentic reinforcement learning}.
\newblock \emph{Preprint}, arXiv:2602.21534.

\bibitem[{Wang et~al.(2024)Wang, Li, Song, Xu, Tang, Zhuge, Pan, Song, Li,
  Singh, Tran, Li, Ma, Zheng, Qian, Shao, Muennighoff, Zhang, Hui, Lin,
  Brennan, Peng, Ji, and Neubig}]{wang2024openhands}
Xingyao Wang, Boxuan Li, Yufan Song, Frank~F. Xu, Xiangru Tang, Mingchen Zhuge,
  Jiayi Pan, Yueqi Song, Bowen Li, Jaskirat Singh, Hoang~H. Tran, Fuqiang Li,
  Ren Ma, Mingzhang Zheng, Bill Qian, Yanjun Shao, Niklas Muennighoff, Yizhe
  Zhang, Binyuan Hui, and 5 others. 2024.
\newblock \href {https://arxiv.org/abs/2407.16741} {{OpenHands: An Open
  Platform for AI Software Developers as Generalist Agents}}.
\newblock \emph{Preprint}, arXiv:2407.16741.

\bibitem[{Wei et~al.(2025)Wei, Duchenne, Copet, Carbonneaux, Zhang, Fried,
  Synnaeve, Singh, and Wang}]{wei2025swerl}
Yuxiang Wei, Olivier Duchenne, Jade Copet, Quentin Carbonneaux, Lingming Zhang,
  Daniel Fried, Gabriel Synnaeve, Rishabh Singh, and Sida~I. Wang. 2025.
\newblock \href {https://doi.org/10.48550/arXiv.2502.18449} {{SWE-RL}:
  Advancing {LLM} reasoning via reinforcement learning on open software
  evolution}.
\newblock \emph{Preprint}, arXiv:2502.18449.
\newblock ArXiv record.

\bibitem[{Wu et~al.(2026{\natexlab{a}})Wu, Lyu, Luo, Wang, Ye, Lin, Ji, Jiang,
  Wang, Wang, Ye, Wang, Wang, Liu, Wang, Cai, Xiong, Fang, Liu, Zhang, Yang,
  Hu, and Liu}]{wu2026eveai4ai}
Kai Wu, Hao Lyu, Zhen Luo, Chaofan Wang, Siyu Ye, Jinghao Lin, Xiaozhong Ji,
  Boyuan Jiang, Shengzhi Wang, Zihan Wang, Yiwen Ye, Hao Wang, Zimu Wang,
  Wenzhe Liu, Ruobing Wang, Kai Cai, Mingliang Xiong, Wen Fang, Mingqing Liu,
  and 4 others. 2026{\natexlab{a}}.
\newblock \href {https://doi.org/10.20944/preprints202608.2108.v1} {{AI4AI}
  survey: From long-horizon agents to recursive self-improvement---definitions,
  reliable horizons, and open problems}.
\newblock Preprints.org.

\bibitem[{Wu et~al.(2026{\natexlab{b}})Wu, Sun, Zhang, Song, Wu, Qi, and
  Cheng}]{wu2026agentstar}
Xixi Wu, Qianguo Sun, Ruiyang Zhang, Chao Song, Junlong Wu, Yiyan Qi, and Hong
  Cheng. 2026{\natexlab{b}}.
\newblock \href {https://arxiv.org/abs/2603.21972} {Demystifying reinforcement
  learning for long-horizon tool-using agents: A comprehensive recipe}.
\newblock \emph{Preprint}, arXiv:2603.21972.

\bibitem[{Xu and Ding(2025)}]{xu2025spo}
Zhongwen Xu and Zihan Ding. 2025.
\newblock \href {https://doi.org/10.48550/arXiv.2509.13232} {{Single-stream
  Policy Optimization}}.
\newblock \emph{Preprint}, arXiv:2509.13232.
\newblock Version v2, 2025-09-23.

\bibitem[{Yang et~al.(2024)Yang, Jimenez, Wettig, Lieret, Yao, Narasimhan, and
  Press}]{yang2024sweagent}
John Yang, Carlos~E. Jimenez, Alexander Wettig, Kilian Lieret, Shunyu Yao,
  Karthik Narasimhan, and Ofir Press. 2024.
\newblock \href {https://arxiv.org/abs/2405.15793} {{SWE-agent}: Agent-computer
  interfaces enable automated software engineering}.
\newblock \emph{Preprint}, arXiv:2405.15793.
\newblock Version 3, 11 November 2024.

\bibitem[{Yang et~al.(2025)Yang, Lieret, Jimenez, Wettig, Khandpur, Zhang, Hui,
  Press, Schmidt, and Yang}]{yang2025swesmith}
John Yang, Kilian Lieret, Carlos~E. Jimenez, Alexander Wettig, Kabir Khandpur,
  Yanzhe Zhang, Binyuan Hui, Ofir Press, Ludwig Schmidt, and Diyi Yang. 2025.
\newblock \href {https://doi.org/10.48550/arXiv.2504.21798} {{SWE}-smith:
  Scaling data for software engineering agents}.
\newblock \emph{Preprint}, arXiv:2504.21798.

\bibitem[{Yao et~al.(2025)Yao, Liu, Zhang, Dong, Shang, and
  Gao}]{yao2025offpolicy}
Feng Yao, Liyuan Liu, Dinghuai Zhang, Chengyu Dong, Jingbo Shang, and Jianfeng
  Gao. 2025.
\newblock \href {https://fengyao.notion.site/off-policy-rl} {Your efficient
  {RL} framework secretly brings you off-policy {RL} training}.
\newblock Technical blog.
\newblock First published August 5, 2025.

\bibitem[{Yu et~al.(2025)Yu, Zhang, Zhu, Yuan, Zuo, Yue, Dai, Fan, Liu, Liu,
  Liu, Lin, Lin, Ma, Sheng, Tong, Zhang, Zhang, Zhang, Zhu, Zhu, Chen, Chen,
  Wang, Yu, Song, Wei, Zhou, Liu, Ma, Zhang, Yan, Qiao, Wu, and
  Wang}]{yu2025dapo}
Qiying Yu, Zheng Zhang, Ruofei Zhu, Yufeng Yuan, Xiaochen Zuo, Yu~Yue, Weinan
  Dai, Tiantian Fan, Gaohong Liu, Lingjun Liu, Xin Liu, Haibin Lin, Zhiqi Lin,
  Bole Ma, Guangming Sheng, Yuxuan Tong, Chi Zhang, Mofan Zhang, Wang Zhang,
  and 16 others. 2025.
\newblock \href {https://doi.org/10.48550/arXiv.2503.14476} {{DAPO: An
  Open-Source LLM Reinforcement Learning System at Scale}}.
\newblock \emph{Preprint}, arXiv:2503.14476.
\newblock Version v2, 2025-05-20.

\bibitem[{Zeng et~al.(2026)Zeng, Huang, Yin, Liu, Li, Li, Wu, Zheng, Zhang,
  Huang, and Qiu}]{zeng2026balanced}
Zhiyuan Zeng, Jiameng Huang, Zhangyue Yin, Jiashuo Liu, Ziniu Li, Bingrui Li,
  Yuhao Wu, Yining Zheng, Ge~Zhang, Wenhao Huang, and Xipeng Qiu. 2026.
\newblock \href {https://doi.org/10.48550/arXiv.2605.04077} {{Balanced
  Aggregation: Understanding and Fixing Aggregation Bias in GRPO}}.
\newblock \emph{Preprint}, arXiv:2605.04077.
\newblock Version v1, 2026-04-14.

\bibitem[{Zheng et~al.(2025)Zheng, Dang, Yu, Li, Jiang, Lin, Liu, Lin, Wu, Hu,
  Yang, Zhou, and Lin}]{zheng2025stabilizing}
Chujie Zheng, Kai Dang, Bowen Yu, Mingze Li, Huiqiang Jiang, Junrong Lin,
  Yuqiong Liu, Hao Lin, Chencan Wu, Feng Hu, An~Yang, Jingren Zhou, and Junyang
  Lin. 2025.
\newblock \href {https://arxiv.org/abs/2512.01374} {{Stabilizing Reinforcement
  Learning with LLMs: Formulation and Practices}}.
\newblock \emph{Preprint}, arXiv:2512.01374.

\end{thebibliography}
\end{document}